\documentclass[lettersize,journal]{IEEEtran}
\usepackage{amsmath,amsfonts}
\usepackage{algorithm}
\usepackage{algorithmicx}
\usepackage[noend]{algpseudocode}
\usepackage{array}
\usepackage[caption=false,font=normalsize,labelfont=sf,textfont=sf]{subfig}
\usepackage{textcomp}
\usepackage{stfloats}
\usepackage{url}
\usepackage{verbatim}
\usepackage{graphicx}
\usepackage{cite}
\usepackage[colorlinks,
            linkcolor=red,   
            anchorcolor=pink,  
            citecolor=green,       
            ]{hyperref}
\usepackage{booktabs}       
\usepackage{nicefrac}       
\usepackage{microtype}      
\usepackage{xcolor}         
\usepackage{makecell}
\usepackage{multirow}
\usepackage{threeparttable}
\usepackage{gensymb}
\usepackage{setspace}
\usepackage{color}
\usepackage{longtable}
\newcommand{\omni}{360{\degree}\ }
\newcommand{\etal}{\textit{et al}.}
\newcommand{\ie}{\textit{i}.\textit{e}.}
\newcommand{\eg}{\textit{e}.\textit{g}.}
\newcommand{\et}{\textit{et\ al.}}

\begin{document}

\title{Perceptual Quality Assessment of \omni Images Based on Generative Scanpath Representation}

\author{Xiangjie Sui*, Hanwei Zhu*, Xuelin Liu, Yuming Fang,~\IEEEmembership{Senior Member, IEEE}, \\Shiqi Wang,~\IEEEmembership{Senior Member, IEEE}, 
Zhou Wang,~\IEEEmembership{Fellow Member, IEEE}
\thanks{
* These authors contributed equally.}
\thanks{
Xiangjie Sui, Xuelin Liu, and Yuming Fang are with the School of Information Management, Jiangxi University of Finance and Economics, Nanchang, Jiangxi, China (e-mail: xjsui@foxmail.com, xuelinliu-bill@foxmail.com, fa0001ng@e.ntu.edu.sg).} \thanks{Hanwei Zhu and Shiqi Wang are with the Department of Computer Science, City University of Hong Kong, Hong Kong, China (e-mail: hanwei.zhu@my.cityu.edu.hk, shiqwang@cityu.edu.hk). }
\thanks{
Zhou Wang is with the Department of Electrical and Computer Engineering, University of Waterloo, Waterloo, Canada (e-mail: zhou.wang@uwaterloo.ca)}
\
}

\markboth{Submitted to IEEE Transactions on Image Processing}%
{Shell \MakeLowercase{\textit{et al.}}: A Sample Article Using IEEEtran.cls for IEEE Journals}


\maketitle
\begin{abstract}
Despite substantial efforts dedicated to the design of heuristic models for omnidirectional (\ie, 360{\degree}) image quality assessment~(OIQA), a conspicuous gap remains due to the lack of consideration for the diversity of viewing behaviors that leads to the varying perceptual quality of \omni images. Two critical aspects underline this oversight: the neglect of viewing conditions that significantly sway user gaze patterns and the overreliance on a single viewport sequence from the \omni image for quality inference.
To address these issues, we introduce a unique generative scanpath representation (GSR) for effective quality inference of \omni images, which aggregates varied perceptual experiences of multi-hypothesis users under a predefined viewing condition.
More specifically, given a viewing condition characterized by the starting point of viewing and exploration time,  a set of scanpaths consisting of dynamic visual fixations can be produced using an apt scanpath generator. 
Following this vein, we use the scanpaths to convert the \omni image into the unique GSR, which provides a global overview of gazed-focused contents derived from scanpaths.
As such, the quality inference of the \omni image is swiftly transformed to that of GSR. 
We then propose an efficient OIQA computational framework by learning the quality maps of GSR.
Comprehensive experimental results validate that the predictions of the proposed framework are highly consistent with human perception in the spatiotemporal domain, especially in the challenging context of locally distorted \omni images under varied viewing conditions. The code will be released at \url{https://github.com/xiangjieSui/GSR}.
\end{abstract}


\section{Introduction}

\IEEEPARstart{V}{irtual} reality (VR) photography endeavors to capture or recreate a spherical natural scene into omnidirectional~(\ie, \omni) images. \omni images offer a vast interactive space, making them particularly appealing in the realm of metaverse applications.  
However, one of the main barriers to the broader adoption of VR photography is the significant loss of quality that \omni images undergo during the processes of image capture, projection, compression, and transmission~\cite{Sun2017_subj, Duan2018_subj}. Consequently, investigating the factors that influence the perceptual quality of \omni images has emerged as a significant research topic.

\begin{figure*}[htp]
\centering
 \includegraphics[width=1\linewidth]{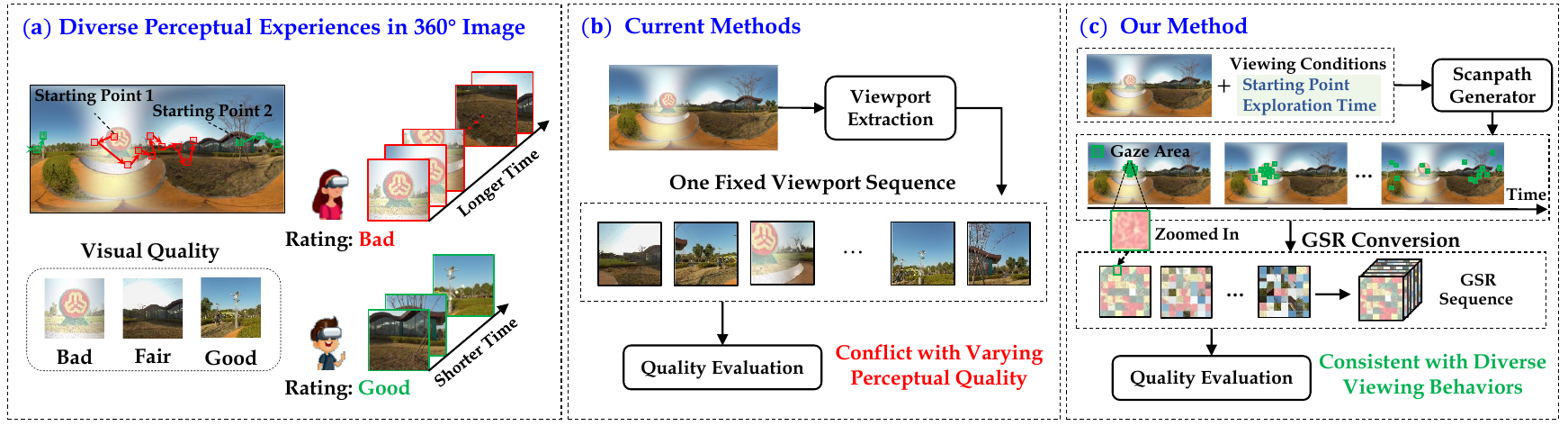}

 \caption{(a) An illustrations of diverse perception experiences in the \omni image. Diverse viewing behaviors lead to varying perceptual quality. (b) The typical scheme of current OIQA methods involves predicting the perceptual quality of \omni images by combining the predicted quality scores of a single fixed viewport sequence. However, the deterministic quality evaluation relying on one fixed sequence cannot explain the probabilistic viewing behaviors with randomness, thereby potentially leading to prediction bias.
 (c) The proposed computational framework. Our method can generate a dynamic GSR sequence given a predefined starting point and an exploration time, and cover multiple sections of the \omni scene at each time instance. Therefore, the predictions of our method can be highly consistent with human perception.}
\label{fig:intro}
\end{figure*}

Researchers have devoted considerable efforts to analyzing the \textit{global} distortions that uniformly affect the quality of \omni images, such as compression, Gaussian blur, and noise~\cite{Sun2017_subj, Duan2018_subj,MC360IQA2019,VGCN2020, MultiStream2021, MFILGN2021, MPBOIQA2021, SPANet2021, Zhang2022, Yang2022, wu2023assessor360}. Numerous subjective user studies have been conducted on degraded \omni images~\cite{Sun2017_subj, Duan2018_subj, SPANet2021}, leading to the development of several objective omnidirectional image quality assessment~(OIQA) models~\cite{MC360IQA2019, VGCN2020, MultiStream2021,  MFILGN2021, SPANet2021, MPBOIQA2021, Zhang2022, Yang2022, wu2023assessor360}. A family of explicit methods aims to extend the 2D full-reference IQA model for the OIQA task~\cite{SPSNR2015,CPPPSNR2016,WSPSNR2017,SSSIM2018} by considering the stretch ratio of the equirectangular projection~\cite{CPPPSNR2016,WSPSNR2017} or the spherical properties of \omni images~\cite{SPSNR2015,SSSIM2018}. However, since comparing entire omnidirectional images is inconsistent with human perception in \omni scenes due to the limited field of view (FoV), a more effective approach is to evaluate the perceptual quality of \omni images by combining quality scores of viewport images~\cite{MC360IQA2019,VGCN2020, MultiStream2021, MFILGN2021,Sui2021,MPBOIQA2021, JUFE22, Zhang2022, Yang2022, wu2023assessor360}. 

Unfortunately, these methods largely overlook the diversity of viewing behaviors that leads to variations in the perceived quality of \omni images, as shown in  Fig.~\ref{fig:intro}~(a). This oversight manifests in two key aspects. 
First, two viewing conditions that significantly
sway user gaze patterns are frequently overlooked. 
Specifically, the \textit{starting point of viewing} and the \textit{exploration time} can significantly impact the scanpath patterns, which in turn affect the perceived quality of \omni images, particularly when images are locally distorted~\cite{Sui2021,JUFE22}. For instance, users may fail to detect distortions if they start viewing from a distortion-free area with a short exploration time. As such, the perceptual quality might vary significantly among users under different viewing conditions.
Second, the majority of OIQA methods predominantly relies on a single fixed viewport sequence from the \omni image to predict its perceptual quality, as shown in Fig.~\ref{fig:intro} (b). 
However, the deterministic quality evaluation relying on one fixed sequence cannot explain the probabilistic viewing behaviors with randomness, thereby potentially leading to prediction bias. 

Despite a few studies have attempted to address these issues~\cite{Sui2021,JUFE22,wu2023assessor360}, they still manifest certain limitations. 
Sui~\etal\cite{Sui2021} proposed an OIQA method by integrating the quality scores of multiple viewport sequences generated using human scanpaths. However, this approach poses significant computational challenges due to the necessity of extracting a multitude of viewport images.
Fang~\etal~\cite{JUFE22} introduced an OIQA model that adapts to different viewing conditions by embedding the positions of the uniformly sampled viewports and the prompts of starting point and exploration time. However, this model did not consider users' viewing behaviors that
can significantly impact the perceptual quality.
Recently, Wu~\etal~\cite{wu2023assessor360} proposed an OIQA model that incorporates a strategy for sampling multiple viewport sequences based on patch-wise entropy. However, such a method may not be adequate to model complex viewing behaviors that are impacted by a variety of factors, \eg, scene semantics and kinematic constraints~\cite{Sitzmann18}.

In this study, we introduce a unique generative scanpath representation (GSR) for OIQA. The underlying principle of the GSR is to aggregate dynamic perceptual experiences of multi-hypothesis users. The GSR conversion is operated under a predefined viewing condition which is characterized by the starting point and exploration time. More specifically, given an arbitrary starting point and exploration time, a set of scanpaths~\cite{scandmm2023} can be produced. This aligns with the philosophy that users' viewing behaviors exhibit significant variation due to the diversity of individual experiences and preferences~\cite{Sui2021}. 
Subsequently, the GSR is constructed by aggregating multiple small gazed-focused areas, as shown in Fig.~\ref{fig:intro} (c). 
As such, the quality inference of \omni image can be swiftly transformed to that of GSR which realistically incorporates the viewing behaviors. Our contributions are summarized as follows: 
\begin{itemize}
     \item We develop a unique GSR representation of \omni images, which is created by using a set of realistic scanpaths generated under a predefined viewing condition. As such, we bridge the gap between viewing conditions and quality assessment in \omni scenes.
     
    \item We provide a comprehensive global overview of dynamic perceptual experiences of multi-hypothesis users with the GSR representation. As such, we can make a thorough quality assessment from a variety of experiences. 

    \item We design a novel computational framework for OIQA based on GSR. Comprehensive experiments demonstrate that our model is much more accurate than advanced OIQA models with less computational complexity.
\end{itemize}


\section{Relate Work}
\begin{table*}[htb]
\centering
\renewcommand\arraystretch{1.2}
\tabcolsep=0.15cm
\small
\caption{OIQA method summary. Trend in OIQA model design tends to introduce scanpaths.}   
    \begin{threeparttable}
    \begin{tabular}{c|lccccl}
    \toprule[0.5mm]
   Type & Method & Year & Category & Scanpath  & Viewing Conditions & Public Availability
    \\ \midrule
    \multirow{4}{*}{\makecell[c]{2D-Plane\\-Based}} 
    
    & CPP-PSNR~\cite{CPPPSNR2016}  & $2016$ & Hand-Craft &  &  &  \url{https://github.com/Samsung/360tools} \\
    
    & WS-PSNR~\cite{WSPSNR2017}  & $2017$ & Hand-Craft &  &  &  \url{https://github.com/Rouen007/WS-PSNR} \\

    & DeepVR-IQA~\cite{Kim2019} & $2019$ & Data-Driven & & & N/A\\ 

    & SAP-Net\cite{SPANet2021} & $2021$ & Data-Driven & & & \url{https://github.com/yanglixiaoshen/SAP-Net}\\ \midrule

    \multirow{2}{*}{\makecell[c]{Sphere\\-Based}} 
    
    & S-PSNR~\cite{SPSNR2015} & $2015$ & Hand-Craft &  &  &  \url{https://github.com/Samsung/360tools}  \\

    & S-SSIM~\cite{SSSIM2018} & $2018$ & Hand-Craft & & & N/A\\ \midrule

    \multirow{8}{*}{\makecell[c]{Viewport\\-Based}}
    & MC360IQA~\cite{MC360IQA2019} & $2019$ & Data-Driven & & & \url{https://github.com/sunwei925/MC360IQA} \\

    & VGCN \cite{VGCN2020} & $2020$ & Data-Driven & & & \url{https://github.com/weizhou-geek/VGCN-PyTorch}\\

    & Sui~\et~\cite{Sui2021} & $2021$ & Hand-Craft & $\checkmark$  & $\checkmark$ & \url{https://github.com/xiangjieSui/img2video}\\

    & Zhou~\et~\cite{MultiStream2021} & $2021$ & Data-Driven & & & N/A\\

    & MFILGN~\cite{MFILGN2021} & $2021$ & Hand-Craft & & & N/A\\

    & MP-BOIQA\cite{MPBOIQA2021} & $2021$ & Hand-Craft & & & N/A\\
    
    & Fang \et\cite{JUFE22} & $2022$ & Data-Driven &  & $\checkmark$ & N/A\\

    & Zhang \et\cite{Zhang2022} & $2022$ & Data-Driven &  & & N/A \\

    & TVFormer~\cite{Yang2022} & $2022$ & Data-Driven& $\checkmark$ & & N/A \\

    & Assessor360~\cite{wu2023assessor360} & $2023$ & Data-Driven & $\checkmark$ & $\checkmark$ & \url{https://github.com/TianheWu/Assessor360}  \\ \midrule

    GSR & GSR-X (Proposed) & 2023 & Data-Driven &  $\checkmark$ &  $\checkmark$ & \url{https://github.com/xiangjieSui/GSR}
    
    \\ \bottomrule[0.5mm]
    \end{tabular}
 
    \end{threeparttable}
    \label{tab:qamd_summary}
\end{table*}

\subsection{Objective Quality Assessment of 2D Images and Videos}
PSNR and SSIM\cite{SSIM2004} are the two most popular full-reference measures for both image and video quality assessment, and they also trigger the development of quality models, including knowledge-driven and data-driven methods~\cite{duanmu2021quantifying,tu2021ugc}. For knowledge-driven methods, the researchers proposed to model the priorities of human visual systems, such as structural similarity~\cite{wang2010information,zeng20123d,DISTS2022}, contrastive sensitive functions~\cite{fang2021superpixel}, free energy theory~\cite{gu2014using,zhai2011psychovisual}, and the information theory~\cite{sheikh2006image,laparra2016perceptual}. Data-driven methods were designed with various feature engineering schemes and learning strategies. Support vector regression (SVR) based on hand-crafted features was one of the most representative methods, aiming to model the natural scene statistics~\cite{bampis2017speed,sheikh2006statistical,mittal2012no,ghadiyaram2017perceptual}. Furthermore, different advanced deep learning techniques have been utilized to regress image or video quality scores, such as convolution neural network (CNN) models~\cite{DBCNN2020,fang2020perceptual,chen2020rirnet,kim2018deep}, transformer-based models~\cite{TreS,zhu2022learning}, meta-learning based model~\cite{zhu2020metaiqa}, learning-to-rank models~\cite{zhang2021uncertainty,zhang2023blind}, and self-supervised learning models~\cite{madhusudana2022image,jiang2022self,ye2012unsupervised}.

\subsection{Objective Quality 
Assessment of Omnidirectional Images}
Current OIQA methods can be roughly classified into three categories based on the planes they operate on -- 2D plane~\cite{CPPPSNR2016, WSPSNR2017, Kim2019, SPANet2021}, sphere~\cite{SPSNR2015, SSSIM2018}, and viewport~\cite{MC360IQA2019,JUFE22,MFILGN2021,VGCN2020, Sui2021, MultiStream2021, Zhang2022, Yang2022, wu2023assessor360}, as listed in Table~\ref{tab:qamd_summary}. The first two categories attempted to extend 2D-IQA methods for OIQA by compensating for the non-uniform
sampling caused by sphere-to-plane projection, \eg, equirectangular projection. Methods in the 2D plane, with typical examples being WS-PSNR~\cite{WSPSNR2017} and CPP-PSNR~\cite{CPPPSNR2016}, weighed local signal errors by their positions. Methods in the sphere, such as S-PSNR~\cite{SPSNR2015} and S-SSIM~\cite{SSSIM2018}, attempted to compute the local quality by uniformly sampling signals on the spherical domain. The last category computed the local quality on the viewport domain. The viewport sampling strategies of these methods can be classified into three categories: predetermined rules~\cite{MC360IQA2019,JUFE22,MFILGN2021,MultiStream2021, MPBOIQA2021}, key points~\cite{VGCN2020, Zhang2022}, and scanpaths~\cite{Sui2021, Yang2022, wu2023assessor360}. As indicated in Table~\ref{tab:qamd_summary}, the trend in the design of the OIQA model tends to involve scanpaths that represent the human viewing behaviors in \omni scenes. However, the effectiveness and practicality of such methods are limited due to the overlook of viewing conditions~\cite{Yang2022}, the reliability of viewing behavior modeling~\cite{wu2023assessor360}, and the time-consuming process of viewport extraction~\cite{Sui2021}. 

\subsection{Scanpath Prediction of Omnidirectional Images}
Scanpath prediction of \omni images aims to produce realistic dynamic gaze behavior based on the human visual perception mechanism. Existing scanpath prediction methods for \omni images can be divided into two categories: saliency-based methods~\cite{SaltiNet2017, Zhu2018, Zhu2020} and generative methods~\cite{Assens2018pathgan, ScanGAN2022,scandmm2023}. Generally speaking, the former first produced the saliency map reflecting the degree of importance of each pixel to the human visual system and then sampled the time-order gaze points from the saliency map to form a scanpath.
The latter took advantage of generative networks, \eg, Generative Adversarial Network (GAN), to directly produce scanpaths based on \omni images. However, such methods did not model the time-dependence of scanpaths and neglected viewing conditions that have a crucial impact on  influencing user viewing behavior~\cite{Sitzmann18, Sui2021, JUFE22}. Recently, Sui~\etal~\cite{scandmm2023} proposed a deep Markov model for scanpath prediction of \omni images, which focused on modeling time-dependent attention in the \omni scenes. More importantly, it was developed with a high degree of interactivity and flexibility, enabling the assignment of specific starting points and exploration times to generate scanpaths.

\section{The Proposed GSR Representation}

\begin{figure*}
\begin{center}
\includegraphics[width=0.98\linewidth]{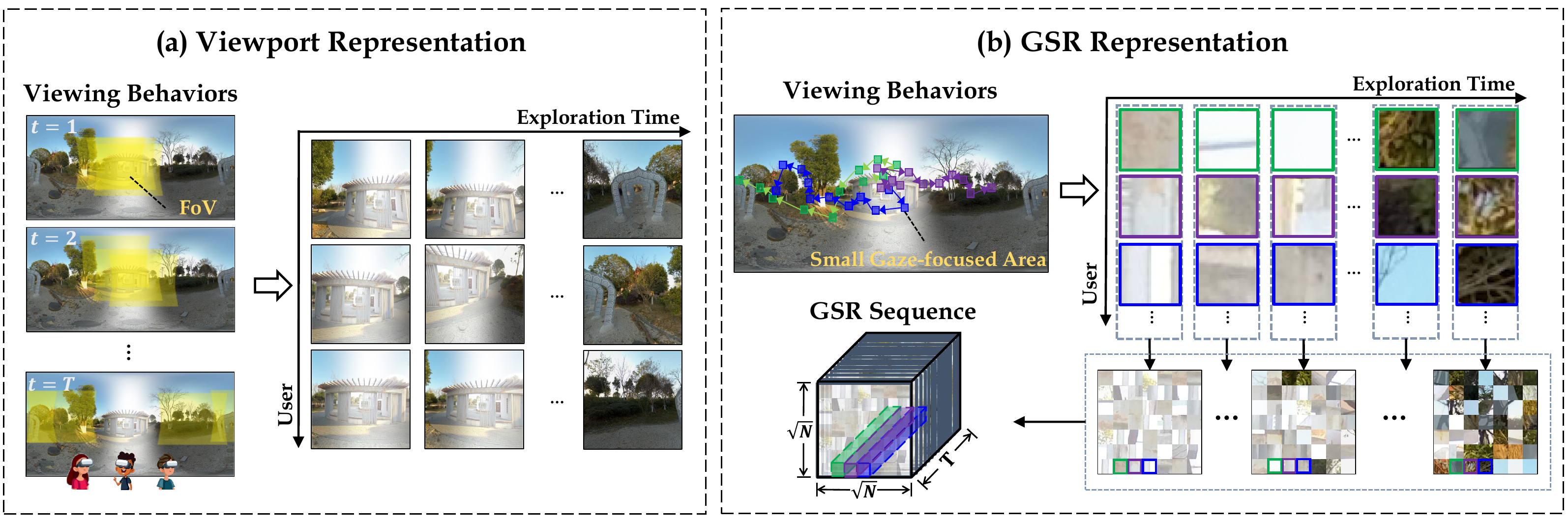}
\end{center}
\vspace{-0.2cm}
\caption{(a) An illustration of viewport representation for \omni images. (b) The proposed GSR representation. The viewports of different users are high likely to overlap. Instead, our GSR representation highlights the small gaze-focused areas and effectively captures the spatio-temporal experiences of multi-hypothesis users. We show a specific example of how we convert a \omni image into a GSR sequence using scanpaths. Note that the patches are extracted in the spherical tangent domain, which we omit for brevity.}

\label{fig:vcr}
\end{figure*}

The GSR aggregates varied perceptual experiences of multi-
hypothesis users under a predefined viewing condition. To this end, the GSR is first designed to adapt to a predefined viewing condition by using a scanpath generator to generate scanpaths based on a given starting point and exploration time. This allows the collection of quality opinions from multi-hypothesis users under a specific viewing condition, similar to the methodology of subjective user studies~\cite{series2012methodology,Sui2021}. To reduce redundant calculations in quality inference for viewport images, which are caused by the high probability of overlap among the FoV of different users (see Fig.~\ref{fig:vcr} (a)), we propose to ``downscaling'' the \omni image to a GSR sequence that realistically incorporates viewing behaviors. Inspired by the visual neuroscience that visual detail is captured primarily by the fovea (\ie, the center of gaze)~\cite{ungerleider2000mechanisms}, the quality inference is conducted on the aggregation of small gaze-focused areas at each time instance (see Fig.~\ref{fig:vcr} (b)). As such, a \omni image is swiftly transformed to a GSR sequence which realistically incorporates viewing behaviors.

\subsection{Scanpath Generator}
We produce realistic scanpaths using an effective scanpath prediction model which is capable of accurately replicating human gaze behaviors. For \omni scenes, the generator should have two essential features: \begin{itemize} 
\item \textit{Adaptability}: it should be able to adapt to any predefined viewing condition, with the capacity to generate scanpaths from any starting point and for any exploration time. 
\item \textit{Generativity}: it should be consistent with probabilistic viewing behaviors, with the capacity of generating different scanpaths for a given \omni image. \end{itemize}
The \textit{adaptability} enables the produced GSR to be relevant and effective in inferring perceptual quality by adapting to predefined viewing conditions. Furthermore, the \textit{generativity} allows for a diverse range of viewing behaviors to be taken into account for a thorough quality inference. An additional benefit of using a generative model is that the resulting scanpaths and corresponding GSR sequence can vary with each iteration, implying that new training samples can be generated regardless of data augmentation~\cite{MAE2022}. The effectiveness of such a strategy is demonstrated in Sec.~\ref{sec:abl}. 

In this study, we use ScanDMM~\cite{scandmm2023} as the scanpath generator within our framework, which takes two inputs: 1) a \omni image $\mathbf{I}$, which is in the form of 2D equirectangular projections; 2) a viewing condition $\mathbf{\Omega}=\{\mathcal{P}_1,\ T\}$ that includes a starting point $\mathcal{P}_1 = (y_1, x_1)$ and an exploration time $T$, where $(y_1, x_1)$ indicates the normalized 2D coordinate at the initial moment with values in the range of $[0, 1]$. Given the two inputs, ScanDMM produces a scanpath through the generative process of the Markov model. This process is carried out in parallel to create $N$ plausible scanpaths: 
\begin{equation}
\begin{split}
\mathcal{\hat{P}}^{1:N}_{1:T} = \mathcal{G}(\mathbf{I}, \mathbf{\Omega}).
\end{split}
\end{equation}
$\mathcal{G}$ is the ScanDMM model. $\mathcal{\hat{P}}^{1:N}_{1:T} = \{\{(y^n_t, x^n_t)\}_{t=1}^{T}\}_{n=1}^{N}$ is a set of $N$ generated scanpaths, where $(y^n_t, x^n_t)$ represents the predicted gaze point of $n$-th hypothetical user at time instant~$t$. 

\subsection{GSR Conversion}
Herein, we detail the process of converting a \omni image into a GSR sequence using the generated scanpaths. Given a predicted gaze position as the center, we extract a mini-patch with a small size of $(\mathbf{P}_h\times \mathbf{P}_w)$ from the image. To account for the overstretch inherent in the equirectangular projection, the patch extraction process is executed using the spherical convolution~\cite{SphereNet2018} which adaptively wraps the kernel around the sphere (see Fig.~\ref{fig:framework} (b)). More specifically, the spherical convolution retrieves the spherical coordinates of sampling points by using a kernel created on the spherical tangent domain. Then, these spherical coordinates are projected to the 2D plane to access the pixel values. 
By repeating the patch extraction process over each hypothetical user, a set of $N$ mini-patches can be obtained at each time instance. These patches are subsequently organized to construct a GSR sequence (denoted as $V_{1:T}$), where each GSR consists of $\sqrt{N}$ rows and columns of patches (see Fig.~\ref{fig:vcr} (b)). Additionally, to capture the temporal quality variation of the independent mini-patch sequence, we impose a constraint on their positions to ensure alignment across time, as suggested in~\cite{wu2022fastquality}. This is done as if an individual mini-patch sequence were a mini-video.

\section{The proposed computational framework}
\begin{figure*}
\begin{center}
\includegraphics[width=0.9\linewidth]{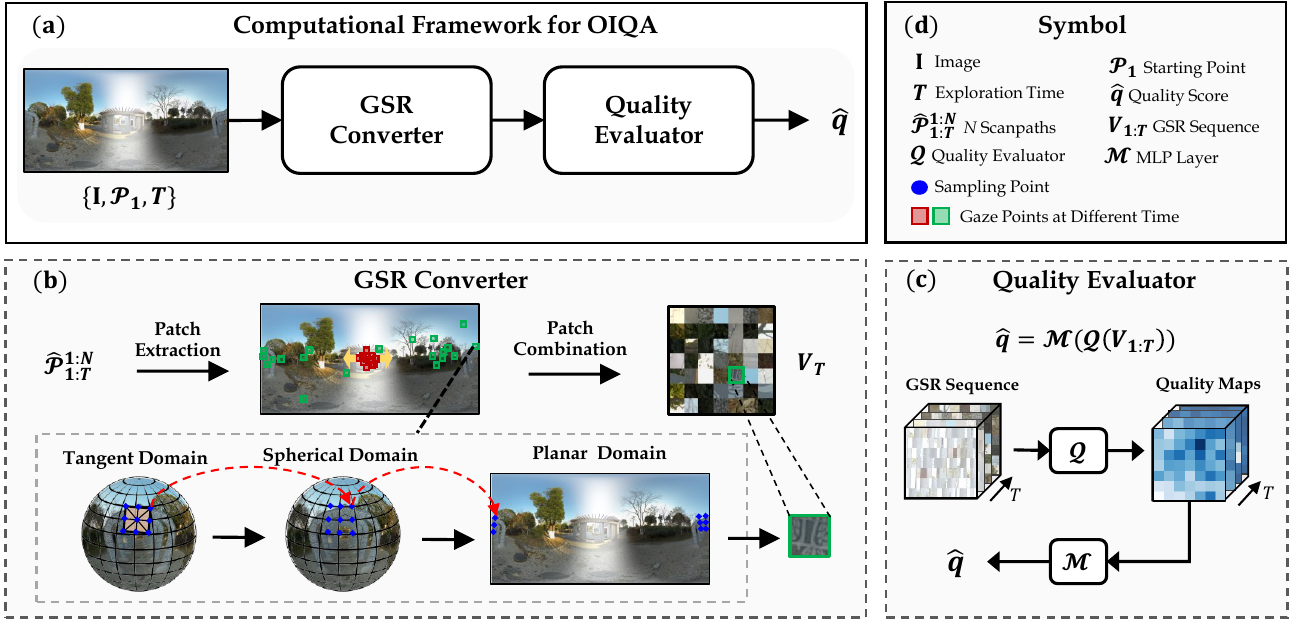}
\end{center}

\caption{Illustrations of the fundamental components for our computational framework. (a) The computational framework in a nutshell. (b) GSR converter. (c) Quality evaluator $\mathcal{M}(\mathcal{Q}(V_{1:T}))$. (d) Symbol description.}
\label{fig:framework}
\end{figure*}

In this section, we provide an overview of the proposed OIQA framework. The framework presented in Fig.~\ref{fig:framework} consists of two components: the GSR converter and the quality evaluator. The former transforms a \omni image into a GSR sequence. The latter extracts quality-aware features from the GSR sequence and regresses them to a quality score. 

Our approach presents \omni images using GSR sequences, which contain essential spatiotemporal information of perceptual experiences for quality inference. 
As such, we leverage an advanced backbone network in video tasks to learn spatiotemporal quality of the GSR sequences. 
Overall, with the resulting GSR sequence of a \omni image, we maximize the following likelihood function:
\begin{equation}
\boldsymbol{\alpha}^* = \arg\max_{\boldsymbol{\alpha}} p(q | \mathcal{M}(\mathcal{Q}(V_{1:T})); \mathbf{\boldsymbol{\alpha}}), 
\end{equation}
where $q$ is the ground-truth quality score and $\boldsymbol{\alpha}$ denotes the learnable parameters in the network.  $\mathcal{Q}$ and $\mathcal{M}$ are the quality evaluator and Multilayer perceptron (MLP) layer, respectively.
The quality inference of the \omni image is achieved by entering the GSR sequence into the quality evaluator and the MLP layer:
\begin{equation}
    \hat{q} = \mathcal{M}(\mathcal{Q}(V_{1:T})),
\end{equation}
where the $\hat{q}$ is the predicted quality score. In this study, we use the X-Clip-B/32~\cite{XCLIP2022} as the quality evaluator. The last MLP layer of X-Clip-B/32 is substituted with the one tailored for video quality assessment~\cite{wu2022fastquality}. We refer to the proposed model as GSR-X. 

\section{Experiments}
In this section, we first offer a detailed description of the implementation of our models and the evaluation procedures. We then compare the proposed model with state-of-the-art quality models. Finally, we present comprehensive ablation studies to analyze the effectiveness of our design elements.

\begin{table*}[]
\centering
\small
\renewcommand\arraystretch{1.35}
\caption{Experimental configurations. The
data in the form of ``\#/\#/\#'' indicates the value of the item in CVIQD/OIQA/JUFE databases.}

    \begin{threeparttable}
    \begin{tabular}{p{2.5cm}p{2.5cm}p{2.5cm}p{2.5cm}p{2.5cm}p{2.5cm}}
    \toprule[0.5mm]
    \multicolumn{6}{c}{\textbf{Default Settings When Viewing Conditions Are Not Accessible}} \\ 
    \multicolumn{6}{c}{Starting point of viewing $\mathcal{P}_1 =(0.5,\ 0.5)$\qquad \quad \qquad Explortion time $T=20$ \qquad \qquad \qquad \qquad}   \\ \midrule
    
    \multicolumn{6}{c}{\textbf{GSR Sequence Configurations}} \\ 
     \multicolumn{6}{c}{\qquad Mini-patch size $(\mathbf{P}_h \times \mathbf{P}_w)$ $=(32\times32)$\qquad \quad  \qquad \ Number of mini-patches (per GSR) $N=49$ \quad \ } 
     \\
    \multicolumn{6}{c}{\quad GSR size $=(224\times224)$\qquad \qquad \qquad \quad Length of the GSR sequence $=T$} \\ \midrule
    
     \multicolumn{6}{c}{\textbf{Specific Model Configurations}} \\ 
 Model & Parameter
     & Learning rate & Batch size & Iteration & Running time* \\ 
    ScanDMM & $18$ MB & $3$e-$4$ & $64$ & $48$ & $2$ \\
    GSR-S & $112$ MB & $3$e-$6$/$3$e-$6$/$1$e-$4$ & $16$/$8$/$16$ & $30$ & $2$/$5$/$25$ \\
    GSR-C & $108$ MB & $3$e-$6$/$3$e-$6$/$1$e-$4$ & $16$/$8$/$16$ & $30$ & $2$/$5$/$22$ \\
    GSR-X & $847$ MB & $8$e-$7$/$8$e-$7$/$8$e-$6$ & $16$/$8$/$16$ & $30$ & $2$/$5$/$22$  
    \\ \midrule
    \multicolumn{6}{l}{Device: AMD Ryzen $9$ $5950$X $16$-Core CPU, $128$ GB RAM, and NVIDIA GeForce RTX 3090 GPU}\\      \bottomrule[0.5mm]
    \end{tabular}
    \end{threeparttable}
    \begin{tablenotes} 
    \item[] * Running time per iteration (minutes).
    \end{tablenotes}
    \label{tab:configurations}
\end{table*}
\begin{table*}[htb]
\renewcommand\arraystretch{1.2}
\centering
\small
\caption{Summary of OIQA Database. The data in the ``\# of images'' column is in the form of ``\# of reference images/\# of distorted images.'' GB: Gaussian blur. GN: Gaussian noise. BD: Brightness discontinuity.}

    \begin{threeparttable}
    \begin{tabular}{lcccccc}
    \toprule[0.5mm]
   Database & Year & Projection &  \#\ of\ Images   &  Resolution  & Viewing Duration &  Distortion\ Types  
    \\ \midrule
     CVIQD~\cite{Sun2017_subj} & $2017$ &  ERP   & $16/528$ & $4$K & - & JPEG,\ AVC,\ HEVC \\
    
     OIQA~\cite{Duan2018_subj} & $2018$ &  ERP  & $16/320$ & $\approx11$K & $20$-second & JPEG,\ JP2K,\ GB,\ GN  \\
    
     JUFE~\cite{JUFE22} & $2022$ &  ERP   & $258/1032$ & $8$K & $5/15$-second & GB,\ GN,\ BD,\ Stitching 
       
       \\ \bottomrule[0.5mm]
    \end{tabular}
    \footnotesize
    \end{threeparttable}
    \label{tab:db_summary}
\end{table*}

\begin{table*}[!htb]
\renewcommand\arraystretch{1.8}
\centering
\small
\nointerlineskip
\caption{Performance comparison. The models in the first and second sections are full-reference and no-reference, respectively. The best performance is highlighted in bold and underlined.}

    \begin{threeparttable}
    \begin{tabular}{l|c|ccc|ccc}
    \toprule[0.4mm]
   \multicolumn{2}{c|}{Database} & CVIQD~\cite{Sun2017_subj} & OIQA~\cite{Duan2018_subj} & JUFE~\cite{JUFE22} & CVIQD~\cite{Sun2017_subj} &  OIQA~\cite{Duan2018_subj} & JUFE~\cite{JUFE22}  
    \\  \hline 
    \multicolumn{2}{c|}{Criterion} & \multicolumn{3}{c|}{SRCC} & \multicolumn{3}{c}{PLCC} \\  \hline
    PSNR & 2D-IQA & $\makecell{0.898 \\ (\textcolor{gray}{\pm0.057})}$ & $\makecell{0.511 \\(\textcolor{gray}{\pm0.046})}$ & $\makecell{0.027 \\ (\textcolor{gray}{\pm0.017})}$ & $\makecell{0.920 \\ (\textcolor{gray}{\pm0.041})}$ & $\makecell{0.565 \\ (\textcolor{gray}{\pm0.076})}$ & $\makecell{0.080 \\ (\textcolor{gray}{\pm0.016})}$ \\ [2ex]
    SSIM~\cite{SSIM2004} & 2D-IQA& $\makecell{0.900 \\ (\textcolor{gray}{\pm0.059})}$ & $\makecell{0.574 \\(\textcolor{gray}{\pm0.058})}$ & $\makecell{0.020 \\ (\textcolor{gray}{\pm0.019})}$	& $\makecell{0.908 \\ (\textcolor{gray}{\pm0.049})}$ &	$\makecell{0.626 \\ (\textcolor{gray}{\pm0.076})}$	& $\makecell{0.058 \\ (\textcolor{gray}{\pm0.035})}$ \\ [2ex]
    DISTS~\cite{DISTS2022} & 2D-IQA& $\makecell{0.934 \\ (\textcolor{gray}{\pm0.016})}$ &	$\makecell{0.907 \\(\textcolor{gray}{\pm0.031})}$ &	$\makecell{0.088 \\ (\textcolor{gray}{\pm0.026})}$ &	$\makecell{0.942 \\ (\textcolor{gray}{\pm0.018})}$ &	$\makecell{0.910 \\(\textcolor{gray}{\pm0.032})}$ &	$\makecell{0.171 \\ (\textcolor{gray}{\pm0.022})}$ \\ [2ex]
    CPP-PSNR~\cite{CPPPSNR2016} & OIQA \& Plane & $\makecell{0.889 \\ (\textcolor{gray}{\pm0.061})}$	& $\makecell{0.510 \\(\textcolor{gray}{\pm0.063})}$	& $\makecell{0.028 \\ (\textcolor{gray}{\pm0.019})}$	& $\makecell{0.909 \\ (\textcolor{gray}{\pm0.050})}$	& $\makecell{0.562 \\(\textcolor{gray}{\pm0.092})}$	& $\makecell{0.073 \\ (\textcolor{gray}{\pm0.015})}$ \\ [2ex]
    WS-PSNR~\cite{WSPSNR2017} & OIQA \& Plane& $\makecell{0.893 \\ (\textcolor{gray}{\pm0.061})}$ &	$\makecell{0.511 \\(\textcolor{gray}{\pm0.066})}$ &	$\makecell{0.026 \\ (\textcolor{gray}{\pm0.018})}$ &	$\makecell{0.912 \\ (\textcolor{gray}{\pm0.048})}$ &	$\makecell{0.560 \\ (\textcolor{gray}{\pm0.101})}$ &	$\makecell{0.077 \\ (\textcolor{gray}{\pm0.015})}$ \\ [2ex]
    S-PSNR~\cite{SPSNR2015} & OIQA \& Sphere& $\makecell{0.893 \\ (\textcolor{gray}{\pm0.061})}$	& $\makecell{0.510 \\(\textcolor{gray}{\pm0.065})}$ &	$\makecell{0.026 \\ (\textcolor{gray}{\pm0.018})}$	& $\makecell{0.913 \\ (\textcolor{gray}{\pm0.047})}$ &	$\makecell{0.560 \\(\textcolor{gray}{\pm0.095})}$ &	$\makecell{0.078 \\ (\textcolor{gray}{\pm0.015})}$ \\ [2ex]
    S-SSIM~\cite{SSSIM2018} & OIQA \& Sphere & $\makecell{0.931 \\ (\textcolor{gray}{\pm0.020})}$ &	$\makecell{0.617 \\(\textcolor{gray}{\pm0.050})}$ &	$\makecell{0.012 \\ (\textcolor{gray}{\pm0.018})}$ &	$\makecell{0.944 \\ (\textcolor{gray}{\pm0.010})}$ &	$\makecell{0.652 \\(\textcolor{gray}{\pm0.061})}$ &	$\makecell{0.055 \\ (\textcolor{gray}{\pm0.016)}}$ \\ [2ex] \hline
    NIQE~\cite{NIQE2013} & 2D-IQA & $\makecell{0.847 \\ (\textcolor{gray}{\pm0.061})}$ & $\makecell{0.702 \\(\textcolor{gray}{\pm0.045})}$ &	$\makecell{0.044 \\ (\textcolor{gray}{\pm0.035})}$ &	$\makecell{0.878 \\ (\textcolor{gray}{\pm0.046})}$ &	$\makecell{0.657 \\ (\textcolor{gray}{\pm0.083})}$ &	$\makecell{0.116 \\ (\textcolor{gray}{\pm0.009})}$ \\ [2ex]
    DBCNN~\cite{DBCNN2020} & 2D-IQA & $\makecell{0.949 \\ (\textcolor{gray}{\pm0.011})}$ & $\makecell{0.927 \\   (\textcolor{gray}{\pm0.032})}$
 & N/A & $\makecell{0.963 \\(\textcolor{gray}{\pm0.009})}$
 & $\makecell{0.931 \\ (\textcolor{gray}{\pm0.027})}$
 & N/A \\ [2ex]

    TreS~\cite{TreS} & 2D-IQA & $\makecell{0.945 \\(\textcolor{gray}{\pm0.013})}$ & $\makecell{0.913 \\(\textcolor{gray}{\pm0.045})}$
 & N/A & $\makecell{0.965 \\(\textcolor{gray}{\pm0.006})}$ & $\makecell{0.918 \\(\textcolor{gray}{\pm0.005})}$
 & N/A \\ [2ex]
    
    MC360IQA~\cite{MC360IQA2019} & OIQA \& Viewport & $\makecell{0.917 \\ (\textcolor{gray}{\pm0.049})}$ & $\makecell{0.900 \\ (\textcolor{gray}{\pm0.031})}$
 & N/A & $\makecell{0.939 \\(\textcolor{gray}{\pm0.036})}$ & $\makecell{0.906 \\ (\textcolor{gray}{\pm0.027})}$
 & N/A \\ [2ex]
    VGCN~\cite{VGCN2020} & OIQA \& Viewport& $\makecell{\mathbf{\underline{0.967}} \\ (\textcolor{gray}{\pm0.006})}$ & $\makecell{\mathbf{\underline{0.955}} \\(\textcolor{gray}{\pm0.027})}$ & N/A &  $\makecell{\mathbf{\underline{0.977}} \\ (\textcolor{gray}{\pm0.002})}$ & $\makecell{\mathbf{\underline{0.964}} \\ (\textcolor{gray}{\pm0.022})}$ & N/A \\ [2ex]
    MFILGN~\cite{MFILGN2021} & OIQA \& Viewport& $\makecell{0.640 \\ (\textcolor{gray}{\pm0.146})}$ &	$\makecell{0.642 \\(\textcolor{gray}{\pm0.149})}$	& $\makecell{0.069 \\ (\textcolor{gray}{\pm0.020})}$	& $\makecell{0.695 \\(\textcolor{gray}{\pm0.128})}$	& $\makecell{0.673 \\ (\textcolor{gray}{\pm0.122})}$ &	$\makecell{0.096 \\(\textcolor{gray}{\pm0.008})}$ \\ [2ex]
    Fang22~\cite{JUFE22} & OIQA \& Viewport& $\makecell{0.908 \\(\textcolor{gray}{\pm0.044})}$ & $\makecell{0.872  \\(\textcolor{gray}{\pm0.048})}$ & $\makecell{0.500 \\(\textcolor{gray}{\pm0.061})}$
 & $\makecell{0.932 \\(\textcolor{gray}{\pm0.033})}$
 & $\makecell{0.879 \\ (\textcolor{gray}{\pm0.054})}$ & $\makecell{0.518 \\(\textcolor{gray}{\pm0.066})}$ \\ [2ex]
 Assessor360~\cite{wu2023assessor360} & OIQA \& Viewport &  $\makecell{0.955 \\(\textcolor{gray}{\pm0.005})}$ & $\makecell{0.946 \\(\textcolor{gray}{\pm0.014})}$ & $\makecell{0.180 \\(\textcolor{gray}{\pm0.043})}$  & $\makecell{0.969 \\(\textcolor{gray}{\pm0.006})}$ & $\makecell{0.953 \\(\textcolor{gray}{\pm0.017})}$ & $\makecell{0.189 \\(\textcolor{gray}{\pm0.048})}$ \\ [2ex]
    GSR-S & OIQA \& GSR & $\makecell{0.905 \\(\textcolor{gray}{\pm0.033})}$ & $\makecell{0.902 \\(\textcolor{gray}{\pm0.019})}$ & $\makecell{0.779 \\(\textcolor{gray}{\pm0.009})}$ & $\makecell{0.937 \\ (\textcolor{gray}{\pm0.020})}$ & $\makecell{0.915 \\(\textcolor{gray}{\pm0.015})}$ & $\makecell{0.789 \\(\textcolor{gray}{\pm0.007})}$\\ [2ex]
    GSR-C & OIQA \& GSR & $\makecell{0.889 \\ (\textcolor{gray}{\pm0.033})}$ & $\makecell{0.845 \\ (\textcolor{gray}{\pm0.023})}$ & $\makecell{0.796 \\ (\textcolor{gray}{\pm0.019})}$ & $\makecell{0.922 \\(\textcolor{gray}{\pm0.028})}$ & $\makecell{0.853 \\ (\textcolor{gray}{\pm0.023})}$ & $\makecell{0.808 \\ (\textcolor{gray}{\pm0.017})}$\\ [2ex]
    GSR-X & OIQA \& GSR & $\makecell{0.944 \\ (\textcolor{gray}{\pm0.019})}$ & $\makecell{0.945 \\ (\textcolor{gray}{\pm0.019})}$ & $\makecell{\mathbf{\underline{0.818}} \\ (\textcolor{gray}{\pm0.022})}$ & $\makecell{0.962 \\ (\textcolor{gray}{\pm0.010})}$ & $\makecell{0.954 \\ (\textcolor{gray}{\pm0.018})}$ & $\makecell{\mathbf{\underline{0.830}} \\ (\textcolor{gray}{\pm0.019})}$\\ 
    \bottomrule[0.4mm]
    \end{tabular}
    \footnotesize
    \end{threeparttable}
    \label{tab:performances}
\end{table*}

\subsection{Implementation Details and Evaluation Protocols}
\label{sec:exp_imp}
The details of the implementation are outlined in Table~\ref{tab:configurations}, and the comparison databases are summarized in Table~\ref{tab:db_summary}.

\vspace{3pt}
\noindent\textbf{Default Settings}. Our framework requires the specification of viewing conditions (\ie, the starting point $\mathcal{P}_1$ and the exploration time $T$) to convert static \omni images into GSR sequences. When such information is not available (\eg, with the CVIQD~\cite{Sun2017_subj} and OIQA~\cite{Duan2018_subj} databases), the conversion is done using default settings. More specifically, the starting point is set to $(0.5,0.5)$, \ie, the center of the \omni images, and the exploration time $T$ is set to $20$ seconds.

\vspace{2pt}
\noindent\textbf{GSR Configurations}. We set the mini-patch size to $32\times32$ and the number of mini-patches per GSR to $49$. This creates a GSR with a size of $224 \times224$, which is a common size for many computer vision networks. The length of a GSR sequence is the same as the exploration time $T$, for example, $20$ in the OIQA database.

\vspace{2pt}
\noindent\textbf{Specific Modules}. We implement the scanpath generator using the original codes provided by the authors~\cite{scandmm2023} and build the quality evaluator using an open source toolbox~\cite{toolbox2022}. The ScanDMM is retrained on the JUFE~\cite{JUFE22} database to learn viewing behaviors in the OIQA task. Note that the scanpath generator and quality evaluator are consistently trained on the same training set to prevent data leakage. 

\vspace{2pt}
\noindent\textbf{Computation Devices}. All experiments are carried out on a server equipped with an AMD Ryzen $9$ $5950$X $16$-Core CPU, a $128$ GB RAM, and an NVIDIA GeForce RTX 3090 GPU. 

\vspace{2pt}
\noindent\textbf{Benchmarking Databases}. We utilize three OIQA databases for model evaluation: CVIQD~\cite{Sun2017_subj}, OIQA~\cite{Duan2018_subj}, and JUFE~\cite{JUFE22}. 
The CVIQD database consists of $528$ distorted \omni images, all with a resolution of $4$K, generated from $16$ distortion-free images with $3$ types of compression distortions at $11$ distortion levels. The OIQA database is a collection of $320$ distorted \omni images with high resolutions ($\approx11$K). These images were created from $16$ reference \omni images, each of which was distorted using $4$ different types of distortion at $5$ different levels. The JUFE database is made up of $1032$ non-uniform distorted \omni images, derived from $258$ reference images, all with a resolution of $8$K. To assess the influence of viewing conditions on the perceptual quality of the \omni images, the subjects were divided into two groups to view a \omni image from two different starting points. They were asked to rate the quality of the \omni image after viewing it for $5$ seconds (while watching) and $15$ seconds (when watching is finished), respectively. This results in a total of $4$ different viewing conditions ($2$ starting points $\times$ $2$ exploration time) in the JUFE database, with each distorted image having $4$ quality labels corresponding to the four viewing conditions. The head and gaze movements data of the subjects have also been recorded to analyze their viewing behaviors.

\vspace{2pt}
\noindent\textbf{Evaluation Protocols}. We use two standard evaluation criteria to quantify quality prediction performance: Spearman's rank-order correlation coefficient (SRCC) and Pearson linear correlation coefficient (PLCC). The higher the SRCC and PLCC values, the better the performance of the model. We randomly divide each database into three sets: training ($70\%$), validation ($10\%$), and test ($20\%$) sets, based on the reference images. We repeat this process five times and report the mean and standard deviation of the SRCC and PLCC results.

\subsection{Performance Comparisons}
In addition to GSR-X model, we have created two other models: GSR-S and GSR-C. These models use Video Swin-T~\cite{SwinT3D2022} and ConvNeXts-T~\cite{ConvNet2022} backbones as the quality evaluator within our proposed framework, respectively~(referring to Table~\ref{tab:configurations} for implementation details). We compare the performance of the proposed models against seven full-reference IQA models, including PSNR, SSIM~\cite{SSIM2004}, DISTS~\cite{DISTS2022}, CPP-PSNR~\cite{CPPPSNR2016}, WS-PSNR~\cite{WSPSNR2017}, S-PSNR~\cite{SPSNR2015}, and S-SSIM~\cite{SSSIM2018}, as well as eight no-reference IQA models, including NIQE~\cite{NIQE2013}, DBCNN~\cite{DBCNN2020}, TreS~\cite{TreS}, MC360IQA~\cite{MC360IQA2019}, VGCN~\cite{VGCN2020}, MFILGN~\cite{MFILGN2021}, Fang22~\cite{JUFE22}, and Assessor360~\cite{wu2023assessor360}. 
When comparing models on the JUFE database, we do not retrain the data-driven models (\eg, VGCN) that overlook the viewing conditions (the results are indicated by ``N/A'' in Table~\ref{tab:performances}) as they suffer from convergence problems when a single \omni image has four quality labels. The results of SRCC and PLCC are summarized in Table~\ref{tab:performances}, from which we make several interesting observations.

\begin{table}[t]
\small 
\renewcommand\arraystretch{1.5}
\tabcolsep=0.4cm
    \centering
    \caption{Performance comparison of different full-reference models. The models prefixed with ``V-'' and ``G-'' denote the models proposed by the study~\cite{Sui2021} and our study, respectively. The suffixes ``AM'' and ``GW'' denote different temporal pooling methods. AM: Arithmetic mean. GW: Ascending half of Gaussian weighting. The best performance of each metric is highlighted.}
    \begin{threeparttable}
    \begin{tabular}{l|cc|c}
    \toprule[0.4mm] 
        Database & \multicolumn{3}{c}{JUFE~\cite{JUFE22}}  
    \\   \hline 
    Criterion & SRCC & PLCC & Time* \\ [0.5ex]  \hline
        PSNR & $\makecell{0.027 \\ (\textcolor{gray}{\pm0.017})}$ & $\makecell{0.080 \\ (\textcolor{gray}{\pm0.016})}$ & -  \\ [2ex]
        SSIM~\cite{SSIM2004} & $\makecell{0.020 \\ (\textcolor{gray}{\pm0.019})}$ & $\makecell{0.058 \\ (\textcolor{gray}{\pm0.035})}$ & - \\ [2ex]
        DISTS~\cite{DISTS2022} & $\makecell{0.088 \\ (\textcolor{gray}{\pm0.026})}$ & $\makecell{0.171 \\ (\textcolor{gray}{\pm0.022})}$ & - \\ [2ex] \hline
        
        V-PSNR-AM & $\makecell{0.551  \\ (\textcolor{gray}{\pm0.021})}$  & $\makecell{0.622 \\ (\textcolor{gray}{\pm0.021})}$ & \multirow{6}{*}{$\makecell{\\ \\ \\ \\ \\127.308}$} \\ [2ex]
        V-PSNR-GW & $\makecell{\mathbf{\underline{0.674}} \\ (\textcolor{gray}{\pm0.021})}$ & $\makecell{\mathbf{\underline{0.679}} \\ (\textcolor{gray}{\pm0.019})}$ &    \\ [2ex]
         V-SSIM-AM & $\makecell{0.578 \\ (\textcolor{gray}{\pm0.026})}$ & $\makecell{0.640 \\ (\textcolor{gray}{\pm0.018})}$ &  \\ [2ex]
        V-SSIM-GW & $\makecell{0.635 \\ (\textcolor{gray}{\pm0.027})}$ & $\makecell{0.648 \\ (\textcolor{gray}{\pm0.019})}$  &  \\ [2ex]
        V-DISTS-AM & $\makecell{0.641 \\ (\textcolor{gray}{\pm0.010})}$ & $\makecell{0.620 \\ (\textcolor{gray}{\pm0.085})}$ &   \\ [2ex]
        V-DISTS-GW & $\makecell{\mathbf{\underline{0.793}} \\ (\textcolor{gray}{\pm0.012})}$ & $\makecell{\mathbf{\underline{0.794}} \\ (\textcolor{gray}{\pm0.011})}$ &  \\ [2ex] \hline
        
        G-PSNR-AM &  $\makecell{0.523 \\ (\textcolor{gray}{\pm0.015})}$ & $\makecell{0.632 \\ (\textcolor{gray}{\pm0.012})}$  & \multirow{6}{*}{$\makecell{\\ \\ \\ \\ \\0.238}$} \\ [2ex]
        G-PSNR-GW & $\makecell{0.541 \\ (\textcolor{gray}{\pm0.018})}$ & $\makecell{0.593 \\ (\textcolor{gray}{\pm0.015})}$  &  \\ [2ex] 
        G-SSIM-AM & $\makecell{0.594 \\  (\textcolor{gray}{\pm0.019})}$ & $\makecell{0.639 \\ (\textcolor{gray}{\pm0.015})}$   &   \\ [2ex]
        G-SSIM-GW & $\makecell{\mathbf{\underline{0.693}} \\ (\textcolor{gray}{\pm0.023})}$ & $\makecell{\mathbf{\underline{0.691}} \\ (\textcolor{gray}{\pm0.021})}$  &  \\ [2ex] 
        
        G-DISTS-AM & $\makecell{0.625 \\ (\textcolor{gray}{\pm0.012})}$ & $\makecell{0.623 \\ (\textcolor{gray}{\pm0.015})}$  &  \\ [2ex]
        G-DISTS-GW & $\makecell{0.770 \\ (\textcolor{gray}{\pm0.019})}$& $\makecell{0.770 \\ (\textcolor{gray}{\pm0.020})}$  &   \\ \bottomrule[0.4mm]
    \end{tabular}
    \end{threeparttable}
    \begin{tablenotes} 
    \item{*} Time cost (seconds) of extracting viewport sequences~\cite{Sui2021} or GSR sequences (our method) for a pair of reference and distorted \omni images.
    \end{tablenotes}
    \label{tab:vcr}
\end{table}

First, even classical 2D-IQA models (\eg, PSNR and SSIM) perform well on CVIQD~\cite{Sun2017_subj} database, indicating the distortion artifacts of this database might be similar to those of traditional 2D-IQA databases. However, MFILGN~\cite{MFILGN2021}, a model based on natural scenes statistics, exhibits lower accuracy and higher variance, possibly due to statistical bias resulting from the limited number of scenes in the training set (\ie, $12$ scenes). On the contrary, the performance of these models drops significantly in the OIQA database~\cite{Duan2018_subj}. This may result from the discrepant distortion artifact procedure -- the distortion artifacts are first injected to each fish-eye image, then stitched and projected onto the 2D-plane. As a result, these distortion artifacts are presented differently from the 2D distortions. 

Second, the results on the two databases with global distortions (\ie, CVIQD and OIQA databases) imply that recent advancements in the field of 2D-IQA are effective in addressing global distortions in \omni images. For example, DISTS~\cite{DISTS2022}, DBCNN~\cite{DBCNN2020}, and TreS~\cite{TreS} show competitive performances compared to the models tailored for \omni images, and outperform all full-reference OIQA metrics (\eg, S-PSNR). This conclusion is further evidenced by the superior results of VGCN~\cite{VGCN2020}, which is built on top of DBCNN and thus benefits from the knowledge regarding 2D distortions. Such observations suggest that, despite the considerable effort put into addressing global distortions in \omni images, they might essentially stand on the same page as 2D-IQA.

\begin{figure*}
    \centering
    \includegraphics[width=0.32\linewidth]{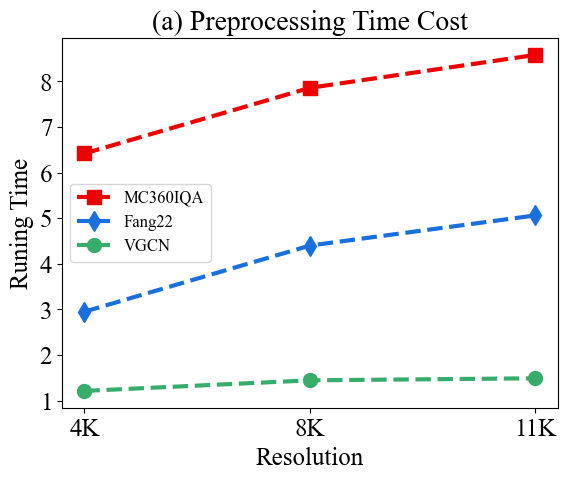}
    \includegraphics[width=0.32\linewidth]{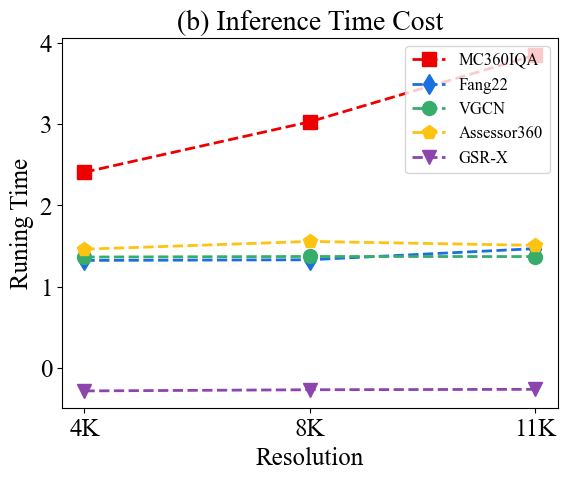}
    \includegraphics[width=0.32\linewidth]{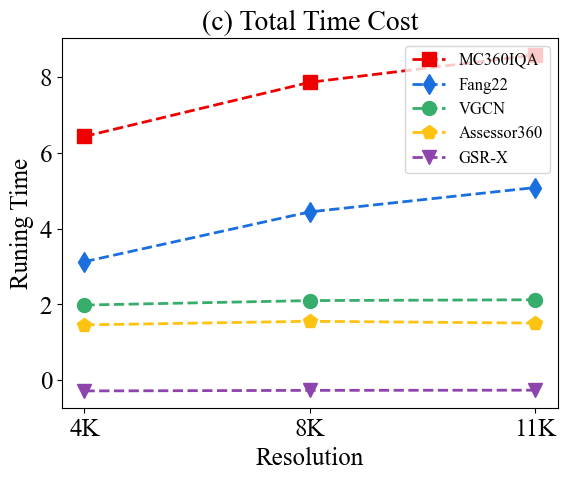}
    \caption{Efficiency comparison in the case of predicting the quality of the \omni image with a resolution of $4$K, $8$K, and $11$K, respectively. We present the $\log_e\hat{t}$ value of the time cost for a better visualization, where $\hat{t}$ represents the time cost in seconds. Our framework operates $\approx0.750$ seconds, regardless of resolution, which is $6$-$7000$ times faster than competing models.}
    \label{fig:time_compar}
\end{figure*}

Third, the results on the database with local distortions (\ie, JUFE database) reveal that current OIQA methods are not able to effectively evaluate the perceptual quality of locally distorted \omni images when viewing conditions are varied. This is evidenced by the very low values of SRCC and PLCC ($\approx0$) of the models. On the contrary, the proposed models achieve superior performance compared to competing models, regardless of the backbone of the quality evaluator, thanks to the flexibility of the scanpath generator and the proposed GSR representation. Furthermore, the proposed OIQA models achieve competitive performance on the CVIQD and OIQA databases, while being much faster than the advanced models~(referring to Fig.~\ref{fig:time_compar}). 

\begin{figure*}
\begin{center}
\includegraphics[width=1\linewidth]{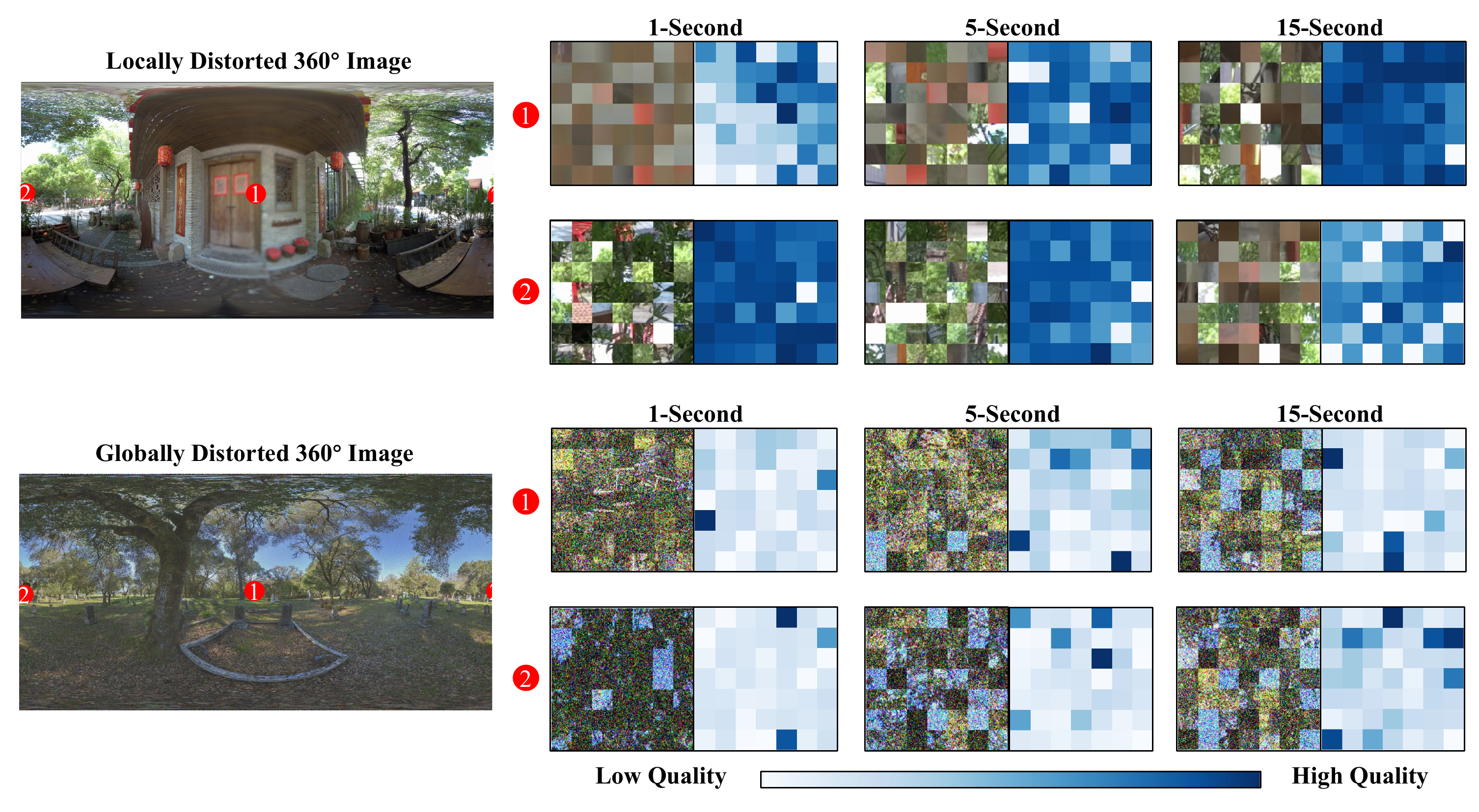}
\end{center}

\caption{The predicted spatiotemporal quality distributions of the GSR sequence when inferring from different starting points of viewing. Top: a locally blurred \omni image (with blur appearing near the point $1$). Bottom: a \omni image affected by global Gaussian noise. }
\label{fig:patch_scores}
\end{figure*}

\subsection{Effectiveness of GSR}\label{sec:exp_vcr}
The GSR is an essential component of the proposed computational framework. Here, we demonstrate the effectiveness of the GSR through quantitative and qualitative experiments.

\vspace{2pt}
\noindent\textbf{Improvement for Full-Reference Metrics.}
We present the performance gains achieved by applying full-reference 2D-IQA methods to GSR sequences, compared to applying them to 2D equirectangular projections~(as baselines) and viewport sequences~\cite{Sui2021}. To ensure a fair comparison, we generate $N=49$ scanpaths to extract viewports. To calculate a quality score for a \omni image using a full-reference metric $Q$, the following process is used:
\begin{equation}
    \hat{q} = \frac{1}{N}\sum^N_{n=1}\frac{\sum^T_{t=1}\omega_t Q(a_{n,t})}{\sum^T_{t=1}\omega_t},
\end{equation}
where $a_{n,t}$ denotes a pair of reference and distorted viewport/GSR images, and $\omega_t$ denotes the weight allocated by the specific temporal pooling method. We add ``V-'' and ``G-'' to the 2D-IQA method as prefixes to name viewport-based models~\cite{Sui2021} and our GSR-based models. In addition, we add suffixes to distinguish different temporal pooling strategies, such as the arithmetic mean~(AM) and the ascending half of Gaussian weighting~(GW)~\cite{Sui2021}.
The final model names follow the format of ``\textit{prefix-Model-suffix}'' (\eg, G-PSNR-GW). The experimental results on the JUFE database are shown in Table~\ref{tab:vcr}. We observe that our methods achieve significant performance improvements compared to baselines and are competitive with the methods proposed by the study~\cite{Sui2021} in terms of performance. Moreover, the proposed GSR-based models run much faster than viewport-based methods by a significant margin.  

\vspace{2pt}
\noindent\textbf{Running Time Comparison.} We speed up quality inference by ``downscale'' a \omni image to a GSR sequence. To show the high computational efficiency of our method, we compare the preprocessing\footnote{Preprocessing refers to the process of extracting viewports. Note that since the Assessor360 model was constructed in an end-to-end management, we only count its inference time.} time, inference time, and total running time of different OIQA models, as depicted in Fig.~\ref{fig:time_compar}. To facilitate visualization, we calculate the $\log_e\hat{t}$ value of the time cost $\hat{t}$ (in seconds). The figures show that the proposed GSR-X model has a stable inference time ($\approx 0.750$ seconds) for a \omni image, regardless of resolution, due to the fixed size of the GSR. This is significantly faster than MCIQA360~\cite{MC360IQA2019} which operates about $1.5$ hours for an $11$ K \omni image. Although VGCN and Assessor360 can maintain stable inference times by compromising image downsampling, their performance is inferior to the GSR-X model (\ie, $\approx7.944$ seconds for VGCN and $\approx4.522$ seconds for Assessor360). In conclusion, our method is $6$-$7000$ times faster than other models. Notably, all the experiments are carried on the same computation devices, with configurations detailed in Sec.~\ref{sec:exp_imp} and Table~\ref{tab:configurations}.

\vspace{2pt}
\noindent\textbf{Qualitative Comparison.} We illustrate that GSR sequences can capture essential spatiotemporal information of hypothetical users' experiences to infer quality by presenting a predicted spatiotemporal quality distribution of the GSR sequences in Fig.~\ref{fig:patch_scores}. This figure shows that when the \omni image is locally distorted, the predicted quality changes gradually in the time domain. For example, when starting to view from point $1$~(referring to the central red circle in Fig.~\ref{fig:patch_scores}), the quality of the initial GSR containing blur distortions is predicted to be low. As the exploration covers more distortion-free regions, the subsequent GSR containing fewer distortions gradually improves in quality. Conversely, for global distortions, the predicted quality is relatively uniform in both the spatial and temporal domains, regardless of the starting points of viewing and exploration time. This observation supports that although GSR has fewer semantics than the raw image, it is capable of capturing essential contents to extract quality-aware features. The resulting quality scores are highly consistent with human judgment, indicating that the proposed models are capable of modeling the dynamic perceptual experiences of humans in the \omni scene.

\begin{table*}[htb]
\small

\renewcommand\arraystretch{1.2}
\centering
\caption{Cross-dataset performance comparison.}

    \begin{threeparttable}
    \begin{tabular}{l|cc|cc}
    \toprule[0.4mm]
   & \multicolumn{2}{c|}{Train on CVIQD~\cite{Sun2017_subj} \& Test on OIQA~\cite{Duan2018_subj}} &  \multicolumn{2}{c}{Train on OIQA~\cite{Duan2018_subj} \& Test on CVIQD~\cite{Sun2017_subj}}
    \\ \hline
    Criterion & SRCC & PLCC & SRCC & PLCC \\ \hline
    MC360IQA~\cite{MC360IQA2019} & \quad $0.288~(\textcolor{gray}{\pm0.070})$ & $0.349~(\textcolor{gray}{\pm0.142})$ & \quad $0.798~(\textcolor{gray}{\pm0.051})$ & $0.842~(\textcolor{gray}{\pm0.044})$ \\
    VGCN~\cite{VGCN2020} & \quad $\mathbf{\underline{0.704}}~(\textcolor{gray}{\pm0.012})$ & $0.710~(\textcolor{gray}{\pm0.012})$ & \quad $0.493~(\textcolor{gray}{\pm0.014})$ & $0.534~(\textcolor{gray}{\pm0.019})$ \\
    Fang22~\cite{JUFE22} & \quad $0.579~(\textcolor{gray}{\pm0.096})$ & $0.615~(\textcolor{gray}{\pm0.086})$ & \quad $0.737~(\textcolor{gray}{\pm0.039})$ & $0.773~(\textcolor{gray}{\pm0.068})$ \\
    Assessor360~\cite{wu2023assessor360} & \quad $0.338~(\textcolor{gray}{\pm0.116})$ & $0.467~(\textcolor{gray}{\pm0.059})$ & \quad $\mathbf{\underline{0.859}}~(\textcolor{gray}{\pm0.010})$ & $\mathbf{\underline{0.893}}~(\textcolor{gray}{\pm0.008})$ \\
    GSR-X & \quad $0.695~(\textcolor{gray}{\pm0.015})$ & $\mathbf{\underline{0.718}}~(\textcolor{gray}{\pm0.014})$ & \quad$0.762~(\textcolor{gray}{\pm0.028})$ & $0.841~(\textcolor{gray}{\pm0.016})$\\
    \bottomrule[0.4mm]
    \end{tabular}
    \footnotesize
    \end{threeparttable}
    \label{tab:cross_dataset}
\end{table*}
\begin{table}[t]
\small
\renewcommand\arraystretch{1.2}
\centering
\tabcolsep=0.075cm

\caption{Ablation studies. The suffix indicates different variants of the GSR-X model. ``w/o $\mathbf{\Omega}$'': assuming the viewing conditions are not accessible. ``w Random'': randomly extracting the mini-patches. ``w Human'': performing the GSR conversion based on human scanpaths. ``w ERP'': cropping the mini-patches on the 2D equirectangular projection plane.
}
\begin{threeparttable}
\begin{tabular}{l|cc|cc}
\toprule[0.5mm]
Database & CVIQD~\cite{Sun2017_subj} &  JUFE~\cite{JUFE22} & CVIQD~\cite{Sun2017_subj} &  JUFE~\cite{JUFE22} 
\\ \hline
Criterion & \multicolumn{2}{c|}{SRCC} & \multicolumn{2}{c}{PLCC} \\ \hline
 GSR-X w/o $\mathbf{\Omega}$ & $0.961$ & $0.140$ & $0.973$ & $0.157$   \\ 
GSR-X w Random & $0.949$ & $0.160$ & $0.962$ & $0.133$ \\ 
 GSR-X w Human & - & $0.810$ & - & $0.816$\\
 GSR-X w ERP & $0.961$ & $0.806$ & $0.972$ & $0.826$ \\
 GSR-X & $\mathbf{\underline{0.961}}$ & $\mathbf{\underline{0.853}}$ & $\mathbf{\underline{0.973}}$ & $\mathbf{\underline{0.859}}$ \\
 \bottomrule[0.5mm]
\end{tabular}
\end{threeparttable}
\label{tab:abla}
\end{table}

\subsection{Generalizability Study}
We compare the generalizability of the proposed GSR-X model with competitive OIQA models in a cross-database setting. We use the CVIQD~\cite{Sun2017_subj} and OIQA~\cite{Duan2018_subj} databases to perform the generalizability comparison. As observed in Table~\ref{tab:cross_dataset}, the models trained on the OIQA database commonly exhibit better generalizability compared to those trained on the CVIQD database. Surprisingly, VGCN~\cite{VGCN2020}, MC360IQA~\cite{MC360IQA2019} and Assessor360~\cite{wu2023assessor360} show a significant performance discrepancy when tested in different cross-database settings. This implies that these models may not be able to effectively extract general quality-aware features and suffer from the overfitting problem. 
On the contrary, although the proposed GSR-X model is based on pre-trained parameters from the video classification task~\cite{XCLIP2022}, which contains a larger domain gap with the quality assessment task, it demonstrates better generalizability in the OIQA task compared with competing models, and is able to adapt to more challenging real-world applications, such as OIQA for locally distorted images \omni under various viewing conditions.

\begin{table}[t]
\small
\renewcommand\arraystretch{1.2}
\centering
\caption{Performance comparison of the GSR-X model using the GSR created with different number of scanpaths.}
\begin{threeparttable}
\begin{tabular}{cc|cc}
\toprule[0.5mm]
$\#$ of Scanpaths & Mini-patch Size & SRCC & PLCC
\\ \hline
$4$ & $112\times 112$ & $0.810$ & $0.819$\\
$16$ & $56\times 56$ & $0.811$ & $0.825$\\
$\mathbf{49}$ & $\mathbf{32\times 32}$ & $\mathbf{\underline{0.853}}$ & $\mathbf{\underline{0.859}}$\\
$64$ & $28\times 28$ & $0.796$ & $0.799$\\
 \bottomrule[0.5mm]
\end{tabular}
\end{threeparttable}

\label{tab:abla_nscanpaths}
\end{table}

\subsection{Ablation Study}
\label{sec:abl}
In this subsection, we analyze the influence of viewing conditions, the scanpath generator, and the spherical tangent representation on the proposed framework by performing ablation experiments. We use the GSR-X as the test model and carry out experiments on the CVIQD~\cite{Sun2017_subj} and JUFE~\cite{JUFE22} databases. The parameters of the comparison models in the same database are maintained to be the same.

\vspace{2pt}
\noindent\textbf{Impact of Accessibility of Viewing Conditions.} We investigate the effect of accessibility of viewing conditions on our method by creating a version of GSR-X that assumes unknown viewing conditions, called ``GSR-X w/o $\mathbf{\Omega}$''. The results shown in Table~\ref{tab:abla} demonstrate that when the \omni images contain local distortions, the accessibility of viewing conditions has a significant impact on the performance of the GSR-X model. This is evidenced by the significant decrease in the performance of ``GSR-X w/o $\mathbf{\Omega}$'' on the JUFE database (\ie, $\approx0.150$ for the PLCC and SRCC values). However, this effect is not observed when the \omni images have global distortions, as evidenced by the satisfactory performance of ``GSR-X w/o $\mathbf{\Omega}$'' on the CVIQD database. These findings are consistent with the study~\cite{Sui2021}, which found that viewing conditions are essential to evaluate the quality of locally distorted \omni images.

\vspace{2pt}
\noindent\textbf{Impact of Scanpath Generator.} We investigate the ability of the scanpath generator to provide diverse viewing behaviors in the \omni scene by asking three questions: 
\begin{itemize}
    \item[1.] Does the temporal information obtained from the scanpath play a critical role in assessing the quality of \omni image?
    \item[2.] Can the scanpath generator be used for data augmentation?
    \item[3.] What is the impact of the number of generated scanpaths on the precision of the predictions?
\end{itemize}
To answer the first two questions, we have created two baseline models: ``GSR-X w Random'' and ``GSR-X w Human''. The former randomly extracts mini-patches from the \omni images to create GSR sequences, thus disregarding the temporal information. The latter uses human scanpaths to create GSR sequences instead of the generative model, meaning that the training samples in each iteration remain the same. The results of the experiment are presented in Table~\ref{tab:abla}, from which two main conclusions can be drawn. First, by comparing the performance of ``GSR-X w Random'' and GSR-X on the two databases, it is evident that temporal information is more significant in evaluating the quality of \omni images that are distorted locally than those that are distorted globally. Second, the proposed GSR-X outperforms ``GSR-X w Human'', indicating that the scanpath generator contributes to data augmentation. This is because new training samples can be generated in each iteration, thus improving the model's performance. Lastly, to investigate the effect of the number of generated scanpaths on prediction accuracy, the GSR-X model was trained using GSR created with $4$, $16$, $49$ and $64$ scanpaths. As the number of scanpaths increases, the mini-patches become smaller with a fixed GSR size (\ie, $224\times224$). As shown in Table~\ref{tab:abla_nscanpaths}, the best performance of the GSR-X model can be achieved when using $49$ scanpaths to create the GSR.

\vspace{2pt}
\noindent\textbf{Impact of Spherical Tangent Representation.} In the GSR conversion, we extract mini-patches from \omni images by utilizing spherical convolution~\cite{SphereNet2018}, which is executed in the spherical tangent domain. Here, we aim to answer the question: 
\begin{itemize}
\item[-] How much performance gains can be achieved by extracting mini-patches in the spherical tangent domain as opposed to cropping mini-patches in the 2D plane?
\end{itemize}
To address this, we construct a baseline model called ``GSR-X w ERP'', which directly crops the mini-patches from a 2D equirectangular projection plane. The results are listed in Table~\ref{tab:abla}, from which we observe that the spherical tangent representation is more effective. This may be attributed to the fact that the spherical tangent representation reduces the geometric distortions of the 2D plane while preserving the spherical properties of \omni images that are not accessible in the 2D plane. For example, the left and right edges of the images with equirectangular projection are discontinuous, which is not the case in the spherical tangent domain. However, the performance gain is less significant on CVIQD database, which may be due to the distortion artifacts being similar to 2D distortion artifacts, as discussed previously.

\section{Conclusion and Discussion}
This paper introduces a novel GSR representation to assess the perceptual quality of \omni images. Our approach involves transforming a static \omni image into a dynamic GSR sequence using a set of scanpaths produced under a specified viewing condition. This representation provides a flexible and effective way to assess the perceptual quality of \omni images under various viewing conditions. The results of our experiments show that the predictions of our framework are in line with human perception in the challenging task of assessing the perceptual quality of locally distorted \omni images under varied viewing conditions.

Our current study focuses on the comprehensive perceptual quality of \omni images under a predefined viewing condition by using a set of generated scanpaths. However, due to the inherent randomness of scanpaths, adaptively selecting those that could improve prediction performance for a given scenario remains an intriguing and challenging problem, yet to be explored. 
Additionally, our framework exhibits potential in \textit{personalized} OIQA, which may be more suitable in VR applications, as user viewing behaviors tend to vary based on individual experiences and preferences. This can be achieved directly by pooling the quality scores of a single patch sequence rather than combining all sequences.

\bibliographystyle{IEEEtran}\
\small{
\bibliography{main}}

\begin{thebibliography}{10}
\providecommand{\url}[1]{#1}
\csname url@samestyle\endcsname
\providecommand{\newblock}{\relax}
\providecommand{\bibinfo}[2]{#2}
\providecommand{\BIBentrySTDinterwordspacing}{\spaceskip=0pt\relax}
\providecommand{\BIBentryALTinterwordstretchfactor}{4}
\providecommand{\BIBentryALTinterwordspacing}{\spaceskip=\fontdimen2\font plus
\BIBentryALTinterwordstretchfactor\fontdimen3\font minus \fontdimen4\font\relax}
\providecommand{\BIBforeignlanguage}[2]{{%
\expandafter\ifx\csname l@#1\endcsname\relax
\typeout{** WARNING: IEEEtran.bst: No hyphenation pattern has been}%
\typeout{** loaded for the language `#1'. Using the pattern for}%
\typeout{** the default language instead.}%
\else
\language=\csname l@#1\endcsname
\fi
#2}}
\providecommand{\BIBdecl}{\relax}
\BIBdecl

\bibitem{Sun2017_subj}
W.~{Sun}, K.~{Gu}, S.~{Ma}, W.~{Zhu}, N.~{Liu}, and G.~{Zhai}, ``A large-scale compressed 360-degree spherical image database: {F}rom subjective quality evaluation to objective model comparison,'' in \emph{IEEE International Workshop on Multimedia Signal Processing}, 2018, pp. 1--6.

\bibitem{Duan2018_subj}
H.~{Duan}, G.~{Zhai}, X.~{Min}, Y.~{Zhu}, Y.~{Fang}, and X.~{Yang}, ``Perceptual quality assessment of omnidirectional images,'' in \emph{IEEE International Symposium on Circuits and Systems}, 2018, pp. 1--5.

\bibitem{MC360IQA2019}
W.~Sun, X.~Min, G.~Zhai, K.~Gu, H.~Duan, and S.~Ma, ``{MC360IQA}: A multi-channel {CNN} for blind 360-degree image quality assessment,'' \emph{IEEE Journal of Selected Topics in Signal Processing}, vol.~14, no.~1, pp. 64--77, 2020.

\bibitem{VGCN2020}
J.~Xu, W.~Zhou, and Z.~Chen, ``Blind omnidirectional image quality assessment with viewport oriented graph convolutional networks,'' \emph{IEEE Transactions on Circuits and Systems for Video Technology}, vol.~31, no.~5, pp. 1724--1737, 2021.

\bibitem{MultiStream2021}
Y.~Zhou, Y.~Sun, L.~Li, K.~Gu, and Y.~Fang, ``Omnidirectional image quality assessment by distortion discrimination assisted multi-stream network,'' \emph{IEEE Transactions on Circuits and Systems for Video Technology}, vol.~32, no.~4, pp. 1767--1777, 2022.

\bibitem{MFILGN2021}
W.~Zhou, J.~Xu, Q.~Jiang, and Z.~Chen, ``No-reference quality assessment for 360-degree images by analysis of multifrequency information and local-global naturalness,'' \emph{IEEE Transactions on Circuits and Systems for Video Technology}, vol.~32, no.~4, pp. 1778--1791, 2022.

\bibitem{MPBOIQA2021}
H.~Jiang, G.~Jiang, M.~Yu, T.~Luo, and H.~Xu, ``Multi-angle projection based blind omnidirectional image quality assessment,'' \emph{IEEE Transactions on Circuits and Systems for Video Technology}, vol.~32, no.~7, pp. 4211--4223, 2022.

\bibitem{SPANet2021}
L.~Yang, M.~Xu, X.~Deng, and B.~Feng, ``Spatial attention-based non-reference perceptual quality prediction network for omnidirectional images,'' in \emph{IEEE International Conference on Multimedia and Expo}, 2021, pp. 1--6.

\bibitem{Zhang2022}
C.~Zhang and S.~Liu, ``No-reference omnidirectional image quality assessment based on joint network,'' in \emph{ACM International Conference on Multimedia}, 2022, p. 943–951.

\bibitem{Yang2022}
L.~Yang, M.~Xu, T.~Liu, L.~Huo, and X.~Gao, ``{TVF}ormer: {T}rajectory-guided visual quality assessment on 360{\degree} images with transformers,'' in \emph{ACM International Conference on Multimedia}, 2022, p. 799–808.

\bibitem{wu2023assessor360}
\BIBentryALTinterwordspacing
T.~Wu, S.~Shi, H.~Cai, M.~Cao, J.~Xiao, Y.~Zheng, and Y.~Yang, ``Assessor360: {M}ulti-sequence network for blind omnidirectional image quality assessment,'' \emph{CoRR}, vol. abs/2305.10983, 2023. [Online]. Available: \url{https://arxiv.org/abs/2305.10983}
\BIBentrySTDinterwordspacing

\bibitem{SPSNR2015}
M.~Yu, H.~Lakshman, and B.~Girod, ``A framework to evaluate omnidirectional video coding schemes,'' in \emph{IEEE International Symposium on Mixed and Augmented Reality}, 2015, pp. 31--36.

\bibitem{CPPPSNR2016}
V.~Zakharchenko, K.~P. Choi, and J.~H. Park, ``{Quality metric for spherical panoramic video},'' in \emph{Optics and Photonics for Information Processing X}, vol. 9970.\hskip 1em plus 0.5em minus 0.4em\relax SPIE, 2016, pp. 57--65.

\bibitem{WSPSNR2017}
Y.~Sun, A.~Lu, and L.~Yu, ``Weighted-to-spherically-uniform quality evaluation for omnidirectional video,'' \emph{IEEE Signal Processing Letters}, vol.~24, no.~9, pp. 1408--1412, 2017.

\bibitem{SSSIM2018}
S.~Chen, Y.~Zhang, Y.~Li, Z.~Chen, and Z.~Wang, ``Spherical structural similarity index for objective omnidirectional video quality assessment,'' in \emph{IEEE International Conference on Multimedia and Expo}, 2018, pp. 1--6.

\bibitem{Sui2021}
X.~Sui, K.~Ma, Y.~Yao, and Y.~Fang, ``Perceptual quality assessment of omnidirectional images as moving camera videos,'' \emph{IEEE Transactions on Visualization and Computer Graphics}, vol.~28, no.~8, pp. 3022--3034, 2022.

\bibitem{JUFE22}
Y.~Fang, L.~Huang, J.~Yan, X.~Liu, and Y.~Liu, ``Perceptual quality assessment of omnidirectional images,'' in \emph{AAAI Conference on Artificial Intelligence}, vol.~36, 2022, pp. 580--588.

\bibitem{Sitzmann18}
V.~Sitzmann, A.~Serrano, A.~Pavel, M.~Agrawala, D.~Gutierrez, B.~Masia, and G.~Wetzstein, ``Saliency in {VR}: How do people explore virtual environments?'' \emph{IEEE Transactions on Visualization and Computer Graphics}, vol.~24, no.~4, pp. 1633--1642, 2018.

\bibitem{scandmm2023}
X.~Sui, Y.~Fang, H.~Zhu, S.~Wang, and Z.~Wang, ``Scan{DMM}: {A} deep markov model of scanpath prediction for 360{\degree} images,'' in \emph{IEEE Conference on Computer Vision and Pattern Recognition}, 2023, pp. 6989--6999.

\bibitem{Kim2019}
H.~G. Kim, H.-T. Lim, and Y.~M. Ro, ``Deep virtual reality image quality assessment with human perception guider for omnidirectional image,'' \emph{IEEE Transactions on Circuits and Systems for Video Technology}, vol.~30, no.~4, pp. 917--928, 2020.

\bibitem{SSIM2004}
Z.~Wang, A.~Bovik, H.~Sheikh, and E.~Simoncelli, ``Image quality assessment: {F}rom error visibility to structural similarity,'' \emph{IEEE Transactions on Image Processing}, vol.~13, no.~4, pp. 600--612, 2004.

\bibitem{duanmu2021quantifying}
Z.~Duanmu, W.~Liu, Z.~Wang, and Z.~Wang, ``Quantifying visual image quality: {A} bayesian view,'' \emph{Annual Review of Vision Science}, vol.~7, no.~1, pp. 437--464, 2021.

\bibitem{tu2021ugc}
Z.~Tu, Y.~Wang, N.~Birkbeck, B.~Adsumilli, and A.~C. Bovik, ``{UGC-VQA}: Benchmarking blind video quality assessment for user generated content,'' \emph{IEEE Transactions on Image Processing}, vol.~30, pp. 4449--4464, 2021.

\bibitem{wang2010information}
Z.~Wang and Q.~Li, ``Information content weighting for perceptual image quality assessment,'' \emph{IEEE Transactions on Image Processing}, vol.~20, no.~5, pp. 1185--1198, 2010.

\bibitem{zeng20123d}
K.~Zeng and Z.~Wang, ``{3D-SSIM} for video quality assessment,'' in \emph{IEEE International Conference on Image Processing}, 2012, pp. 621--624.

\bibitem{DISTS2022}
K.~Ding, K.~Ma, S.~Wang, and E.~P. Simoncelli, ``Image quality assessment: Unifying structure and texture similarity,'' \emph{IEEE Transactions on Pattern Analysis and Machine Intelligence}, vol.~44, no.~5, pp. 2567--2581, 2022.

\bibitem{fang2021superpixel}
Y.~Fang, Y.~Zeng, W.~Jiang, H.~Zhu, and J.~Yan, ``Superpixel-based quality assessment of multi-exposure image fusion for both static and dynamic scenes,'' \emph{IEEE Transactions on Image Processing}, vol.~30, pp. 2526--2537, 2021.

\bibitem{gu2014using}
K.~Gu, G.~Zhai, X.~Yang, and W.~Zhang, ``Using free energy principle for blind image quality assessment,'' \emph{IEEE Transactions on Multimedia}, vol.~17, no.~1, pp. 50--63, 2014.

\bibitem{zhai2011psychovisual}
G.~Zhai, X.~Wu, X.~Yang, W.~Lin, and W.~Zhang, ``A psychovisual quality metric in free-energy principle,'' \emph{IEEE Transactions on Image Processing}, vol.~21, no.~1, pp. 41--52, 2011.

\bibitem{sheikh2006image}
H.~R. Sheikh and A.~C. Bovik, ``Image information and visual quality,'' \emph{IEEE Transactions on Image Processing}, vol.~15, no.~2, pp. 430--444, 2006.

\bibitem{laparra2016perceptual}
V.~Laparra, J.~Ball{\'e}, A.~Berardino, and E.~P. Simoncelli, ``Perceptual image quality assessment using a normalized laplacian pyramid,'' \emph{Electronic Imaging}, vol. 2016, no.~16, pp. 1--6, 2016.

\bibitem{bampis2017speed}
C.~G. Bampis, P.~Gupta, R.~Soundararajan, and A.~C. Bovik, ``Sp{EED-QA}: {S}patial efficient entropic differencing for image and video quality,'' \emph{IEEE Signal Processing Letters}, vol.~24, no.~9, pp. 1333--1337, 2017.

\bibitem{sheikh2006statistical}
H.~R. Sheikh, M.~F. Sabir, and A.~C. Bovik, ``A statistical evaluation of recent full reference image quality assessment algorithms,'' \emph{IEEE Transactions on Image Processing}, vol.~15, no.~11, pp. 3440--3451, 2006.

\bibitem{mittal2012no}
A.~Mittal, A.~K. Moorthy, and A.~C. Bovik, ``No-reference image quality assessment in the spatial domain,'' \emph{IEEE Transactions on Image Processing}, vol.~21, no.~12, pp. 4695--4708, 2012.

\bibitem{ghadiyaram2017perceptual}
D.~Ghadiyaram and A.~C. Bovik, ``Perceptual quality prediction on authentically distorted images using a bag of features approach,'' \emph{Journal of Vision}, vol.~17, no.~1, pp. 32, 1--25, 2017.

\bibitem{DBCNN2020}
W.~Zhang, K.~Ma, J.~Yan, D.~Deng, and Z.~Wang, ``Blind image quality assessment using a deep bilinear convolutional neural network,'' \emph{IEEE Transactions on Circuits and Systems for Video Technology}, vol.~30, no.~1, pp. 36--47, 2020.

\bibitem{fang2020perceptual}
Y.~Fang, H.~Zhu, Y.~Zeng, K.~Ma, and Z.~Wang, ``Perceptual quality assessment of smartphone photography,'' in \emph{IEEE Conference on Computer Vision and Pattern Recognition}, 2020, pp. 3677--3686.

\bibitem{chen2020rirnet}
P.~Chen, L.~Li, L.~Ma, J.~Wu, and G.~Shi, ``{RIRN}et: {R}ecurrent-in-recurrent network for video quality assessment,'' in \emph{ACM International Conference on Multimedia}, 2020, pp. 834--842.

\bibitem{kim2018deep}
W.~Kim, J.~Kim, S.~Ahn, J.~Kim, and S.~Lee, ``Deep video quality assessor: {F}rom spatio-temporal visual sensitivity to a convolutional neural aggregation network,'' in \emph{European Conference on Computer Vision}, 2018, pp. 219--234.

\bibitem{TreS}
S.~A. Golestaneh, S.~Dadsetan, and K.~M. Kitani, ``No-reference image quality assessment via transformers, relative ranking, and self-consistency,'' in \emph{IEEE Winter Conference on Applications of Computer Vision}, 2022, pp. 3209--3218.

\bibitem{zhu2022learning}
H.~Zhu, B.~Chen, L.~Zhu, and S.~Wang, ``Learning spatiotemporal interactions for user-generated video quality assessment,'' \emph{IEEE Transactions on Circuits and Systems for Video Technology}, vol.~33, no.~3, pp. 1031--1042, 2023.

\bibitem{zhu2020metaiqa}
H.~Zhu, L.~Li, J.~Wu, W.~Dong, and G.~Shi, ``{MetaIQA}: {D}eep meta-learning for no-reference image quality assessment,'' in \emph{IEEE Conference on Computer Vision and Pattern Recognition}, 2020, pp. 14\,143--14\,152.

\bibitem{zhang2021uncertainty}
W.~Zhang, K.~Ma, G.~Zhai, and X.~Yang, ``Uncertainty-aware blind image quality assessment in the laboratory and wild,'' \emph{IEEE Transactions on Image Processing}, vol.~30, pp. 3474--3486, 2021.

\bibitem{zhang2023blind}
W.~Zhang, G.~Zhai, Y.~Wei, X.~Yang, and K.~Ma, ``Blind image quality assessment via vision-language correspondence: A multitask learning perspective,'' in \emph{IEEE Conference on Computer Vision and Pattern Recognition}, 2023, pp. 14\,071--14\,081.

\bibitem{madhusudana2022image}
P.~C. Madhusudana, N.~Birkbeck, Y.~Wang, B.~Adsumilli, and A.~C. Bovik, ``Image quality assessment using contrastive learning,'' \emph{IEEE Transactions on Image Processing}, vol.~31, pp. 4149--4161, 2022.

\bibitem{jiang2022self}
S.~Jiang, Q.~Sang, Z.~Hu, and L.~Liu, ``Self-supervised representation learning for video quality assessment,'' \emph{IEEE Transactions on Broadcasting}, vol.~69, no.~1, pp. 118--129, 2022.

\bibitem{ye2012unsupervised}
P.~Ye, J.~Kumar, L.~Kang, and D.~Doermann, ``Unsupervised feature learning framework for no-reference image quality assessment,'' in \emph{IEEE Conference on Computer Vision and Pattern Recognition}, 2012, pp. 1098--1105.

\bibitem{SaltiNet2017}
M.~Assens, X.~Giro-i Nieto, K.~McGuinness, and N.~E. O’Connor, ``Salti{N}et: Scan-path prediction on 360 degree images using saliency volumes,'' in \emph{IEEE International Conference on Computer Vision Workshops}, 2017, pp. 2331--2338.

\bibitem{Zhu2018}
Y.~Zhu, G.~Zhai, and X.~Min, ``The prediction of head and eye movement for 360 degree images,'' \emph{Signal Processing: Image Communication}, vol.~69, pp. 15--25, 2018.

\bibitem{Zhu2020}
Y.~Zhu, G.~Zhai, X.~Min, and J.~Zhou, ``The prediction of saliency map for head and eye movements in 360 degree images,'' \emph{IEEE Transactions on Multimedia}, vol.~22, no.~9, pp. 2331--2344, 2020.

\bibitem{Assens2018pathgan}
A.~Marc, G.-i.-N. Xavier, M.~Kevin, and E.~O. Noel, ``Path{GAN}: {V}isual scanpath prediction with generative adversarial networks,'' in \emph{European Conference on Computer Vision Workshops}, 2018.

\bibitem{ScanGAN2022}
D.~Martin, A.~Serrano, A.~W. Bergman, G.~Wetzstein, and B.~Masia, ``Scan{GAN}360: {A} generative model of realistic scanpaths for 360{\degree} images,'' \emph{IEEE Transactions on Visualization and Computer Graphics}, vol.~28, no.~5, pp. 2003--2013, 2022.

\bibitem{series2012methodology}
B.~Series, ``Methodology for the subjective assessment of the quality of television pictures,'' \emph{Recommendation ITU-R BT}, pp. 500--13, 2012.

\bibitem{ungerleider2000mechanisms}
S.~K. Ungerleider and L.~G, ``Mechanisms of visual attention in the human cortex,'' \emph{Annual Review of Neuroscience}, vol.~23, no.~1, pp. 315--341, 2000.

\bibitem{MAE2022}
K.~He, X.~Chen, S.~Xie, Y.~Li, P.~Doll\'ar, and R.~Girshick, ``Masked autoencoders are scalable vision learners,'' in \emph{IEEE Conference on Computer Vision and Pattern Recognition}, 2022, pp. 16\,000--16\,009.

\bibitem{SphereNet2018}
B.~Coors, A.~P. Condurache, and A.~Geiger, ``Sphere{N}et: {L}earning spherical representations for detection and classification in omnidirectional images,'' in \emph{European Conference on Computer Vision}, 2018, pp. 518--533.

\bibitem{wu2022fastquality}
H.~Wu, C.~Chen, J.~Hou, L.~Liao, A.~Wang, W.~Sun, Q.~Yan, and W.~Lin, ``{FAST}-{VQA}: {E}fficient end-to-end video quality assessment with fragment sampling,'' in \emph{European Conference of Computer Vision}, 2022, pp. 538--554.

\bibitem{XCLIP2022}
B.~Ni, H.~Peng, M.~Chen, S.~Zhang, G.~Meng, J.~Fu, S.~Xiang, and H.~Ling, ``Expanding language-image pretrained models for general video recognition,'' in \emph{European Conference of Computer Vision}, 2022, pp. 1--18.

\bibitem{NIQE2013}
A.~Mittal, R.~Soundararajan, and A.~C. Bovik, ``Making a {``}completely blind{''} image quality analyzer,'' \emph{IEEE Signal Processing Letters}, vol.~20, no.~3, pp. 209--212, 2013.

\bibitem{toolbox2022}
\BIBentryALTinterwordspacing
H.~Wu, ``Open source deep end-to-end video quality assessment toolbox,'' 2022. [Online]. Available: \url{http://github.com/timothyhtimothy/fast-vqa}
\BIBentrySTDinterwordspacing

\bibitem{SwinT3D2022}
Z.~Liu, J.~Ning, Y.~Cao, Y.~Wei, Z.~Zhang, S.~Lin, and H.~Hu, ``Video swin transformer,'' in \emph{IEEE Conference on Computer Vision and Pattern Recognition}, 2022, pp. 3202--3211.

\bibitem{ConvNet2022}
Z.~Liu, H.~Mao, C.-Y. Wu, C.~Feichtenhofer, T.~Darrell, and S.~Xie, ``A {C}onv{N}et for the 2020s,'' in \emph{IEEE Conference on Computer Vision and Pattern Recognition}, 2022, pp. 11\,966--11\,976.

\end{thebibliography}
\end{document}